\documentclass[runningheads,orivec,final]{llncs}
\usepackage{microtype}
\usepackage[T1]{fontenc}    
\usepackage[utf8]{inputenc} 
\usepackage{xspace}   
\usepackage[disable]{todonotes}    
\usepackage{graphicx}   
\usepackage[hyphens]{url}
\usepackage{breakurl}
\usepackage[hyperindex,breaklinks,hidelinks]{hyperref}
\usepackage{cleveref}

\usepackage{url}

\newcommand\eg{e.\,g.\xspace}
\newcommand\ie{i.\,e.\xspace}

\newcommand\wrt{w.\,r.\,t.\xspace}

\newboolean{showcomments}
\setboolean{showcomments}{true} 
\ifthenelse{\boolean{showcomments}}
{\newcommand{\nb}[2]{
  \fcolorbox{black}{yellow}{\bfseries\sffamily\scriptsize#1}
  {\sf\small$\blacktriangleright$\textit{#2}$\blacktriangleleft$}
 }
 
}
{\newcommand{\nb}[2]{}
 
}
\usepackage[xcolor=orange]{changes}
\definechangesauthor[name={Claudio Menghi},color=orange]{CM}
\definechangesauthor[name={Maximilian Köhl},color=blue]{MK}

\newtheorem{characterization}{Characterization}

\begin{document}
\title{Towards a Characterization \\ of Explainable Systems\thanks{This work is heavily based on discussions and work presented during the GI-Dagstuhl Seminar 19023 \emph{Explainable Software for Cyber-Physical Systems} (ES4CPS) \cite{seminar}. We wish to thank all participants for the fruitful and valuable discussions, especially Francisco Javier Chiyah Garcia, Claudio Menghi, and Andreas Wortmann.}}
\author{Dimitri Bohlender\inst{1}\and
Maximilian A. Köhl\inst{2}
}
\institute{RWTH Aachen University, Aachen, Germany\and
Saarland University, Saarland Informatics Campus, Saarbrücken, Germany}
\maketitle 

\begin{abstract}
  Building software-driven systems that are easily understood becomes a challenge, with their ever-increasing complexity and autonomy.
  Accordingly, recent research efforts strive to aid in designing \emph{explainable systems}.
  Nevertheless, a common notion of what it takes for a system to be explainable is still missing.
  To address this problem, we propose a characterization of explainable systems that consolidates existing research.
  By providing a unified terminology, we lay a basis for the classification of both existing and future research, and the formulation of precise requirements towards such systems.
\end{abstract}

\section{Introduction}
The desire to sufficiently understand systems we interact with is natural.
If a person acts in an unexpected way, \eg comes late to an important appointment, one might ask for an explanation and be content with learning about a traffic jam.
In contrast, software-driven systems become more and more opaque with their ever-increasing complexity and autonomy.
This trend culminates when even engineers and experts are unable to understand them.
Consequently, recent research revolving around \emph{explainable systems} aims to provide \emph{appropriate} means to make complex systems comprehensible by users and engineers alike.

The reasons for building explainable systems are numerous.
For instance, explanations allow engineers to locate and fix bugs,
and enable effective cooperation.
Leaving users in the dark about the inner workings of a system does not only make it less trustworthy, potentially causing distrust, \cite{baum2017:two} but can also lead to drastic operating errors \cite{plane:crash}.
Therefore, systems should be designed with explainability in mind, especially if they impact human lives.

Although it is already difficult to explain
purely software-based systems, the uncertain and non-linear nature of cyber-physical systems aggravates the problem even further.
For some systems, such as autonomous vehicles with machine learning based components,
it is often not even clear what kind of explanation to consider appropriate.
Certainly, many existing approaches in the area of explainable artificial intelligence (XAI) do produce explanations \cite{SHAP,Strumbelj:2014}.
However, those explanations are technical and intended for experts, rendering them inappropriate for laypersons.
To evaluate different approaches
a common foundation consolidating existing work is required.

To this end, we propose a characterization of \emph{explainable systems}.
Our characterization provides a common terminology which allows us to classify, compare, and evaluate existing research.
It clarifies how different research in the area is connected.
The provided terminology is a starting point towards establishing precise domain-specific explainability requirements.
We do not aim to provide technical definitions of explainability but rather a general conceptual framework centered around the characterization.
For concrete application domains, this characterization needs to be instantiated by technical definitions.

\paragraph{Outline.}
This paper is structured as follows: aiming at a characterization of what it means for a system to be explainable we start by characterizing the prior notion of \emph{explanation}.
We evaluate our characterization of explanations along its dimensions and address potential criticism.
Subsequently, we lift the characterization of explanations to characterize \emph{explainable systems}.
We conclude this paper by classifying and comparing existing research in the field using our terminology, and by pointing out further research directions.


\section{Explanations}
\label{sec:explanation}
Prima facie, what makes a system explainable are explanations of some kind.
So, what is an explanation?
Explanations come in a multitude of different forms.
Smoke rising over a house can be an explanation for hearing sirens, quantum physics explains the results of the double-slit experiment, and a desire not to be robbed presumably explains why someone locks their doors.
For our purposes, however, we do neither aim to give a comprehensive characterization capturing those occurrences of explanations, nor do we have to.
Instead, we aim to provide a characterization of explanations in the context of software-driven systems---we are interested in explaining aspects of such systems.

In the context of software-driven systems, explanations are desired because there is some lack of understanding of certain aspects of a system by someone.
An explanation of those aspects is supposed to resolve this lack of understanding.
The reasons for wanting to understand are diverse.
For instance, understanding may foster trust of the end-user in a system or allow an engineer to locate and fix a bug.
These considerations make it plausible that there is no one \emph{explanation of a system} as a whole.
Instead, an explanation explains some aspect of a system, we call the \emph{explanandum} \cite{sep:scientific-explanation}, to someone or a broader group of people, we call the \emph{target group}.
These observations give rise to the following characterization of explanations in the context software-driven systems:

\begin{characterization}[Explanation]
\label{def:explanation}
A representation $E$ of some information $I$ is an \emph{explanation} of explanandum $X$ with respect to target group $G$ iff the processing of $E$ by any representative agent $A$ of $G$ makes $A$ understand $X$.
\end{characterization}

In what follows, we discuss various facets of this characterization in more detail and argue for its adequacy and usefulness on the basis of the following running example.

\paragraph{Running Example.}
We consider an automotive navigation system entrusted to plan the temporally shortest route to the driver's home.
However, at some point, the computed route leads off the freeway just to reenter it again at the next possibility, effectively bypassing a section of the freeway on a more speed-limited road.
This route may seem implausible to the driver and begs the question why the navigation system does not favor staying on the freeway.

\subsection{Information and Representation}
\label{sec:inforepr}
Let's first address the distinction between the information $I$ and its representation $E$.
Intuitively, the same information can be represented in multiple different ways.
The roadway system of a city, for instance, can be represented graphically or in textual form.
Representations of both kinds, if they are adequate, carry information \emph{about} which streets are connected to which other streets. 
They carry this information, because, there is a causal relationship between the roadway system and those representations---if the streets would be connected in another way, then those representations would be different.\footnote{We use a simple counterfactual analysis of causation \cite{sep:causation-counterfactual} here. However, this is not crucial for our analysis. One may use another theory of causation.}
For a representation to be adequate, it needs to be produced by a system ensuring that it carries the relevant information.
Information in that sense\footnote{\label{fo:shannon}This notion of information goes back to Shannon \cite{shannon:information}. Interestingly, information theory also gives us a lower bound on the size of an explanation given that it needs to carry the information relevant for understanding.} is \emph{factive}, i.e., there is no room for misinformation.
In contrast, representations can misrepresent \cite{Neander:2017}.

What sets apart faithful explanations from made up stories is that faithful explanations do not misrepresent but actually carry relevant information about the explanandum.
That is, they do not carry some arbitrary information but rather the information necessary for understanding and this information lines up with their representational content.
Separating information and representation enables us to draw this crucial distinction.

Further, the factive notion of information is \emph{agent-independent}.
Therefore, it allows us to evaluate different representations carrying the \emph{same} information.
In the context of our example, imagine that the reason for the seemingly implausible routing is the system's reaction to a message about an accident on the freeway, received via the Traffic Message Channel (TMC).
The TMC message log, however, will be too low-level and verbose to be an explanation for the average human driver.
While it contains the relevant information, namely that there is an accident on the otherwise temporally shorter route, it represents it in a way accessible to an engineer but not to the average human driver.
In contrast, an indication of the accident's location with a symbol on the map might explain the navigation system's reasoning sufficiently to a driver.
Clearly, presenting information in different forms leads to explanations with different qualities.

Not all parts of a representation carry information.
For example, streets could be represented by red or blue lines without the color carrying any information about the roadway system, \ie, the color does not reflect any property of the roadway system but is a mere design choice.
Again, other parts of a representation may carry information not relevant for understanding.
The type of a street, for instance, indicated by a symbol on a map, is irrelevant for understanding why a route is the spatially shortest route.
It is essential that the representation carries at least the information relevant for understanding.

While the information carried by a representation is agent-independent, its representational meaning is not.
What is a representation of what depends on the representation consuming, producing, or processing system.
A bit string, for instance, may represent a multitude of different things depending on how it is produced, consumed, or used in a computation.
For the proposed characterization the representation is processed by agents of the target group.
Therefore, what is represented by $E$ is determined by the representative agents.

\subsection{Processing and Understanding}
\label{sec:procund}
A central idea of the proposed characterization is that the \emph{processing} of  explanation $E$ is what makes agent $A$ \emph{understand} the explanandum $X$.

Processing, in general, is a very broad and vague term, but it fits our purposes as, analogously, explanations may take all kinds of different forms.
For example, an explanation may be provided as a visual or acoustic stimulus.
Within those more broad categories, there are again subcategories.
A visual representation, for instance, could be a map or a text.
Processing is not even bound to happen consciously.
In fact, most processing within the human mind happens unconsciously \cite{kihlstrom1987cognitive}.
Concretizing what \emph{processing} amounts to will only be possible after fixing both the form an explanation takes and the target group.
In the case of human agents, it is advisable to turn to psychology and the cognitive sciences to get a grip on how humans process representations of different form.

In any case, if an agent is unable to process an alleged explanation, then this alleged explanation does not explain anything to her and therefore shall not count as an explanation with respect to the agent.
The definiens of the characterization presupposes that the agent is indeed able to process the explanation.
Accordingly, something\footnote{Depending on how one understands \emph{representation}, being a representation already implies that it is processible by any representative agent of the target group.} only counts as an explanation with respect to a target group if any representative agent of that group is able to process it.

In the context of the navigation system example: if the explanation is only given in acoustic form, it will not be processable by a deaf person, and therefore not qualify as an explanation with respect to her.
Whether it counts as an explanation with respect to a target group depends on whether the ability to hear is required for an agent to be representative of the group.

Processing an explanation requires the use of cognitive or computational resources.
In practice, further constraints could be tied to \emph{processing}, like an upper bound on the time it may take, which, in turn, will constrain what counts as an explanation under constraints.
To this end, the characterization could be extended with contexts.
Intuitively, a complicated matter may be explainable in some context but not in another, if the time is constraint.\footnote{Referring to footnote \ref{fo:shannon}, the lower bound on the size of an explanation also gives us a lower bound on the time needed to process it completely.}

The mere ability to process a representation, of course, does not make this representation an explanation.
It is important, that the processing makes any representative agent understand the explanandum.
One may argue that this characterization merely is a shift in terminology which is not very illuminating with regard to what an explanation is.
Per se, this objection is not a legitimate objection against the characterization but instead demands to clarify further what understanding amounts to.
While it is true that the characterization shifts the elusiveness from \emph{explanation} to \emph{understanding} the latter concept is much more tangible.
As computer scientists, we have an intuition of what it means to understand a system.
Further, there is existing work in psychology and the cognitive sciences \cite{trabasso2003story,keil2006explanation} concerned with understanding which can be leveraged to get a grip on the concept of understanding in our context.

Relying on the concept of understanding, our characterization allows us to separate different concerns.
If we are the target group, then we are the authority to judge what makes us understand and, hence, according to our characterization, counts as an explanation of what.
Depending on the target group, the authority of what it means to understand falls into different hands, and so does the authority of what counts as an explanation.
For instance, for the end-user, one may want to consult psychology and the cognitive sciences in order to get a grip on what it means for an end-user to understand some aspect of a system. 

Clarifying what understanding means is an explicit part of concretizing the characterization for an application domain and most likely involves further interdisciplinary studies with the respective target group.
For our purposes, aiming at a general characterization involving different target groups, we, therefore, rely on some general notion of \emph{understanding} which, admittedly is vague, but does not inadequately anticipate the outcome of further necessary interdisciplinary research.
Part of designing an explainable system is to make sure that alleged explanations make the relevant target group understand the explanandum, that is, that they are indeed explanations of the explanandum according to our characterization, and not something else.
By relying on the concept of understanding, our characterization makes this need explicit instead of hiding it.

Another crucial feature of the proposed characterization is that it is not sufficient for an agent to merely think that she understands the explanandum after processing an alleged explanation.
While it is hard to spell this out precisely, there certainly is a difference between genuine understanding and the mere feeling that one understood something.
One idea to measure whether an agent actually understands the explanandum is to measure the effects of an explanation.
For instance: is an engineer able to locate and fix a bug after processing an explanation for it?
In general, one should exercise caution, when trying to measure understanding in terms of such effects of an explanation. 
Just providing an engineer with the fix for a bug may not necessarily make her understand the bug despite potentially enabling her to fix it.\footnote{Thanks to Jan Reineke for pointing that out.}

Existing research in the area of explainable systems is very broad.
Nevertheless, this research rightly claims to provide explanations and make systems explainable.
The link between those lines of work is not precise and technical but instead rooted in the vague concept of understanding.
Again, our characterization makes this link explicit by relying on the concept of understanding.

\subsection{Explanandum}
The need for an explanation originates in a lack of understanding.
An explanation is always an explanation \emph{for}\ or \emph{of}\footnote{While an explanation \emph{for} usually comprises reasons for events, an explanation \emph{of} may comprise reasons but may also just provide further details necessary for understanding. 
We use \emph{explanation of} as the more general version.} something that is explained.
In the philosophy of science the phenomenon which is explained by an explanation is called the \emph{explanandum} \cite{sep:scientific-explanation}. We borrow this term for our purposes.

A lack of understanding may concern some event happened within a system, a decision of a system, the behavior of a system under certain circumstances, the question how to achieve a specific goal by using a system, or some other aspect, quality, or property of a system.
For instance, in the context of our running example,
the lack of understanding may be described by the question: ``why does the fastest route not continue over the freeway?''

Naturally, when communicating a lack of understanding it needs to be described in some way.
Currently and in the example above, the primary system to describe a lack of understanding is natural language.
As natural language is imprecise, what precisely the is meant by the above question may not be clear.
The necessity of managing this imprecision is driving a lot of research in the requirements engineering community \cite{NLReqsAmbiguous,NLRE}.
Again, our characterization does not hide this but makes it explicit.
Future research needs to focus on how to describe a lack of understanding precisely.
For existing research \cite{SHAP} there usually exist explananda in the form of technical descriptions and definitions.

There neither is a unique explanation of an explanandum nor does an explanation, in general, explain a unique explanandum.
This enables comparisons of different explanations of the same explanandum across different dimensions.
By concretizing what understanding amounts to and providing a common description language for explananda, within an application domain, different approaches can be compared with regard to the explanations they provide.

\subsection{Target Group and Representative Agent}
In \Cref{sec:inforepr}, we argued that the message log in our running example does not qualify as an explanation for the average human driver while it may serve as an explanation to an engineer.
Referring to target groups, like average human drivers, introduces imprecision because it is rarely possible to specify precise characteristics and skills of a certain group \cite{UserCentredRE}.
Nevertheless, it enables generalization.
For our purposes, we aim to be able to express when something is explainable \wrt a particular group.
A characterization which does not generalize and abstract away from every concrete individual will not be very useful.

Still, a target group $G$ may contain an agent who lacks the required abilities and knowledge.
Referring to our running example, there may be a driver who is inexperienced with reading maps or does not understand a certain symbol indicating an accident.
To avoid such corner cases causing a representation $E$ to not qualify as an explanation, even though it explains the aspect of interest to the average human driver, our characterization refers to \emph{representative} agents.
Such representative agents are equipped with the background knowledge and processing capabilities characteristic for the target group.
This is a common approach to this kind of problem and can also be found in requirements engineering where users with specific skills are assumed to reason about a system \cite{sutcliffe2012user}.


\section{Explainable Systems}
In the previous section, we introduced a characterization of \emph{explanation} and argued for its feasibility in the context of software-driven systems.
In this section, we lift this characterization in order to specify what it takes for a system to be explainable.
Intuitively, what makes a system explainable \wrt a target group is access of the group to explanations as laid out in \Cref{sec:explanation}.

To provide access to explanations they need to be produced by something or someone. This \emph{means for producing} an explanation could be another system, the system itself, or even a human expert.
The mere theoretical existence of some explanation is, however, not sufficient for a system to be explainable.
These considerations give rise to the following characterization:

\begin{characterization}[Explainable System]
  A system $S$ is \emph{explainable} by means $M$ with respect to some explanandum\footnote{Here $X$ is not an arbitrary explanandum but an explanandum related to $S$.} $X$ and target group $G$ iff $M$ is able to produces an $E$ such that $E$ is an explanation of $X$ \wrt $G$.
\end{characterization}

Just as the notion of \emph{explanation} is not absolute but rather relative to a target group and an explanandum, a system is not just explainable but rather explainable with respect to certain aspects and a target group.
We envision that in the long-run, domain-specific requirements of what needs to be explainable to whom will emerge.
Calling a system simply \emph{explainable} may then implicitly refer to a clear standard of what is explainable to whom.

\subsection{System}
When talking about systems, we are thinking of a delineated set of entities which may interact to realize some functionality as a whole.
In the context of software-driven systems, some entities will be purely software-based.
However, the characterization is not specific to a certain kind of system but applicable to a wide range of systems, ranging from models to hardware-encompassing cyber-physical systems.
In particular, we do not restrict ourselves to functions that model the behavior of a system.
A system may be a concrete implementation that is to be explained with respect to the function it implements.


\subsection{Means for Producing an Explanation}
In general, an explanation of some aspect $X$ of $S$ may be provided by something that or somebody who is detached from $S$.
Reconsidering our navigation system example, the driver might wonder what the system's capabilities are, for instance, whether she can enforce that the route continues on the freeway.
Such information about a system is already known at design time and static.
As a result, the typical explanation of what a system can and cannot do is provided by human engineers as a manual.
Manuals explain how to use a system in order to archive certain goals.
They are usually provided by humans and perfectly fit within our characterization.
Typically manuals are also targeted at a specific group.

If a system $S$ itself is the means of explanation for some aspect $X$ of itself \wrt a target group $G$, \ie, if $M = S$, we call this system \emph{self-explainable}. Self-explainable systems are just a special case of our characterization.

\begin{characterization}[Self-Explainability]
  A system $S$ is \emph{self-explainable} with respect to some explanandum $X$ and target group $G$ iff $S$ is explainable by $S$ with respect to explanandum $X$ and target group $G$.
\end{characterization}

While there are no conceptual constraints on $M$, the system's application domain might have requirements towards it.
Just as there may be additional constraints on the processing of an explanation, \ie, the consumption side, there may be additional constraints on the production side of the explanation as well.
For example, a realtime scenario might introduce an upper bound on the time the means may take to produce an explanation or security policies may require that the produced explanation does not leak specific information.

If there is no appropriate means $M$ which would be able to produce an explanation $E$ of $X$ \wrt $G$, \eg, for computability reasons, then the aspect of interest $X$ is just impossible to explain.
Analogously to computability theory, not everything in explainable---not even in theory.
Even if something is explainable in theory it may still be an enormous practical challenge to actually design an appropriate means $M$ producing the explanations.

Concerning the characterization of \emph{explainable systems} one may argue that it would be insufficient if $M$ would just be able to produce an $E$ but does not actually do so when necessary, \eg, if an agent demands an explanation.
We kept this complication out of the characterization for now.
As already pointed out in \Cref{sec:procund} we plan to extend the characterization with contexts in future work.
Introducing contexts would enable further requirements towards $M$ concerning in which contexts which explanations need to be produced.

As argued in \Cref{sec:inforepr} a faithful explanation $E$ needs to carry the relevant information and must not misrepresent.
To this end, $M$ needs to be a reliable source for information about $X$.
Presumably, it needs some kind of deep access into the internals of the system $S$ it provides explanations for.
Based on the extracted information it then needs to produce the representation $E$.

\todo{A section on terminology.}

\section{Charting the Field}
Research on explainable systems is highly interdisciplinary. In this section, we categorize some disciplines based on the proposed characterization.

With regard to explainable systems, research in the areas of human-machine interaction, natural-language generation, cognitive sciences, and psychology focuses on representations $E$.
The primary research question is: how can information be represented such that it is easily understood by a representative human agent?
This involves studies on how the relevant target group processes the respective representations.
Further, work on conversational interfaces is to be classified as work on representations and the interaction with $M$.

Within computer science, the area of requirements-engineering is concerned with the process of defining and maintaining requirements \cite{kotonya1998requirements}.
We propose to treat explainability as a special kind of non-function requirement.
We believe that requirements engineering shall focus on (A) how to specify explainability as a non-functional requirement and (B) how to specify requirements towards explanations and means of producing explanations.
Results should be methodologies and guidelines on how to ensure that a system is indeed explainable.

For some systems under development, explainability can be achieved constructively.
Being concerned with the design and implementation of software, software-engineering is the natural home for research on how to design systems such that they are \emph{explainable by design}.
Research in artificial intelligence suggests that the belief–desire–intention (BDI) software model~\cite{BDH+01}---developed for engineering intelligent agents---lends itself to make a software agent's beliefs, desires and intentions explicit. Here, beliefs are assumptions over the agent's world and intentions are decisions the agent has taken to achieve its desires (goals). 
Making these explicit and accessible (e.g., through logging, modeling, or knowledge representation techniques) can support developers and users at system runtime to explain the behavior of (multi-)agent systems. 
For some programming languages, logging frameworks (e.g., ELIOT\footnote{ELIOT logging framework: \url{https://eliot.readthedocs.io/en/latest/}}) are available that support structured logging to provide more accessible abstractions of system runtime behavior.
Other research focuses on how to achieve \emph{explainability by design} by utilizing argumentation-theory based reasoning to make decisions \cite{EthicsToExplainability}.

Within the area of artificial intelligence, there is both, research on how to explain more traditional AI systems, such as expert systems \cite{swartout1991explanations}, and work on how to explain more recently used techniques, such as deep neural networks \cite{SHAP}.
The latter is known as the field of \emph{eXplainable Artificial Intelligence} (XAI).
As most modern machine-learning techniques are opaque by nature, the research here focuses on how to extract relevant and meaningful information from such systems that does explain the aspects of interest.

Formal languages are of interest to precisely specify requirements towards explanations and means of producing.
Further, explanations may be fleshed out as formal objects that can be investigated and represent the relevant information, in contrast to natural language, in a precise manner.
In the area of verification, one may want to investigate how to verify that systems are explainable or that explanations are appropriate, \eg, sufficiently accurate.

\section{Conclusion}
Software-driven systems become more and more opaque and incomprehensible with their ever-increasing complexity and autonomy.
To address this problem, recent research efforts focus on making systems explainable.
Nevertheless, a common foundation of what it means for a system to explainable is still missing.
As a common foundation, we propose a characterization of \emph{explainable systems} which connects the different lines of research claiming to work on explainable systems.
We illustrate the application of the characterization and its terminology on the example of an automotive navigation system.

We believe that a common terminology fosters the systematic discussion of topics in the area of explainable systems.
It further allows comparing and evaluating different lines of work in terms of the dimensions of the characterization, such as the target group and the representations they provide.

From the perspective of requirements engineering, our characterization of explainable systems is a non-functional requirement.
Given the recent interest in building explainable systems, we envision that guidelines and methodologies for building them, \eg, meta-models of explainable systems, will emerge.
The proposed characterization is a first step towards having precise domain-specific certifications which attest certain levels of explainability.

A central result of our investigation is that explainability is not an absolute requirement but relative.
A system is not just explainable \emph{per se} but only with respect to certain aspects and to certain groups.

In future work, we want to extend the characterization with contexts and study qualities of explanations such as accuracy or their effects.

\bibliographystyle{splncs04}
\bibliography{literature}

\end{document}